\documentclass[sigconf]{acmart}
\newcommand{\modified}[1]{\textcolor{black}{#1}}
\AtBeginDocument{%
  \providecommand\BibTeX{{%
    \normalfont B\kern-0.5em{\scshape i\kern-0.25em b}\kern-0.8em\TeX}}}

\copyrightyear{2024}
\acmYear{2024}
\setcopyright{acmlicensed}\acmConference[HRI '24]{Proceedings of the 2024 ACM/IEEE International Conference on Human-Robot Interaction}{March 11--14, 2024}{Boulder, CO, USA}
\acmBooktitle{Proceedings of the 2024 ACM/IEEE International Conference on Human-Robot Interaction (HRI '24), March 11--14, 2024, Boulder, CO, USA}
\acmDOI{10.1145/3610977.3634983}
\acmISBN{979-8-4007-0322-5/24/03}

\usepackage{microtype,stfloats}
\acmSubmissionID{1116}

\begin{document}

\title{Sprout: Designing Expressivity for Robots Using Fiber-Embedded Actuator}

\author{Amy Koike}
\orcid{0009-0000-9088-6578}
\affiliation{
  \institution{University of Wisconsin--Madison}
  \city{Madison}
  \state{Wisconsin}
  \country{USA}
  \postcode{53706}
}
\email{ekoike@wisc.edu}

\author{Michael Wehner}
\orcid{0000-0001-8423-7870}
\affiliation{%
  \institution{University of Wisconsin--Madison}
  \city{Madison}
  \state{Wisconsin}
  \country{USA}
  \postcode{53706}
  }
\email{wehner2@wisc.edu}

\author{Bilge Mutlu}
\orcid{0000-0002-9456-1495}
\affiliation{%
  \institution{University of Wisconsin--Madison}
  \city{Madison}
  \state{Wisconsin}
  \country{USA}
  \postcode{53706}
  }
\email{bilge@cs.wisc.edu}

\renewcommand{\shortauthors}{Amy Koike, Michael Wehner, \& Bilge Mutlu}

\begin{abstract}
In this paper, we explore how techniques from soft robotics can help create a new form of robot expression. We present \textit{Sprout}, a soft expressive robot that conveys its internal states by changing its body shape. \textit{Sprout} can extend, bend, twist, and expand using \textit{fiber-embedded actuators} integrated into its construction. These deformations enable \textit{Sprout} to express its internal states, for example, by expanding to express anger and bending its body sideways to express curiosity. Through two user studies, we investigated how users interpreted \textit{Sprout}'s expressions, their perceptions of \textit{Sprout}, and their expectations from future iterations of \textit{Sprout}'s design. We argue that the use of soft actuators opens a novel design space for robot expressions to convey internal states, emotions, and intent.
\end{abstract}

\ccsdesc[300]{Human-centered computing~Interface design prototyping}

\ccsdesc[500]{Computer systems organization~Robotics}

\keywords{Human-robot interaction, soft robotics, nonverbal behavior, expressivity, fiber-embedded actuator}

\begin{teaserfigure}
  \includegraphics[width=\columnwidth]{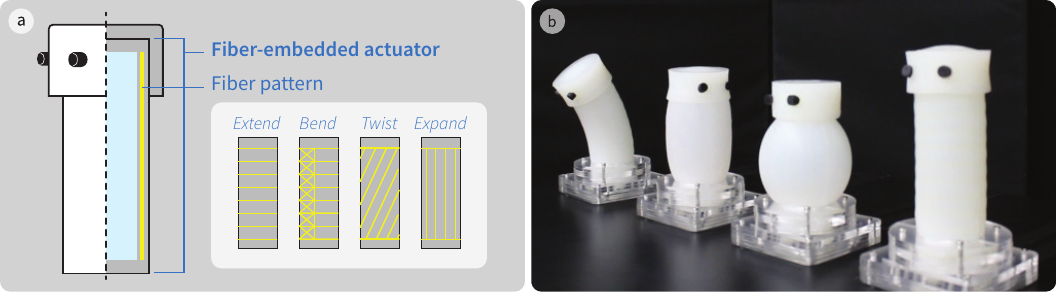}
  \caption{In this study, we explore how techniques from soft robotics, specifically \textit{fiber-embedded actuators} (a) might be used to design robots, such as \textit{Sprout} (b) that can express emotions and internal states by changing the shape of their bodies.}
  \Description{An illustration is on the left. A photo is on the right. The illustration on the left includes a robot interior breakdown, which is a schematic design. It also includes visual representations for four fiber embedding patterns. The photo on the right shows four robots aligned, and deforming differently.}
  \label{fig:teaser}
\end{teaserfigure}

\maketitle

\section{Introduction}


    Robots are designed to communicate with their users with the aid of several mechanisms for expression. 
    Prior research in human-robot interaction (HRI) showed that mirroring human expressive capabilities, including the use of narrative gestures \cite{huang2013modeling}, facial expressions \cite{breazeal1999build}, and nodding \cite{liu2012generation}, can create and enrich communication between humans and robots. 
    However, this approach largely assumes a humanlike form in the design of the robot's body and may not be appropriate or effective in creating expressions for non-humanlike robot designs. 
    To address the communication requirements of the diverse gallery of robot designs, including human-like, animal-like, abstract, and so on, it is important to explore new forms of expression beyond humanlike expressions.
    Additionally, the vast majority of HRI studies have been undertaken with rigid robots, comprised of rigid links actuated about discrete joints using an array of electric motors, which may constrain the creation of new forms of expression for robots.
    To explore this broader design space for robot expressivity, this work leverages \textit{soft robotics}, specifically recent developments in soft pneumatic actuators \cite{walker2020soft}. 
    

    A \textit{soft pneumatic actuator (SPA)} is a type of soft robot typically built of elastomeric chambers that deform in one or more directions when inflated with pressurized gas.
    Because of their advantages in high compliance, safety, and large deformation, SPAs have found a wide range of applications, such as bio-medical devices \cite{ranzani2015bioinspired}, grippers \cite{zaidi2021actuation}, or haptic feedback devices \cite{tawk2019soft}. 
    Besides these practical purposes, several works utilized SPAs for new robot expressions, \textit{e.g.,} to develop robot skin \cite{Hu2020} or robotic muscles \cite{larsen2022wisard} using SPAs, toward enhancing human-robot interactions.
    These preliminary studies have demonstrated the promise of soft robotics for novel robot social cues, yet the use of soft robotics principles and methods in the design and construction of social robots is still underexplored. 

    To explore the use of soft robotics in designing robot expressivity, we designed and fabricated \textit{Sprout}, a social robot that conveys its emotions and internal states by changing the shape of its body (Figure \ref{fig:teaser}).
    The \textit{Sprout}'s body is made of a \textit{fiber-embedded actuator} that allows \textit{Sprout} to deform to create expressions. 
    The \textit{fiber-embedded actuator} is a type of SPA reinforced by fibers, and the fiber pattern creates various deformation modes upon inflation.
    This approach was particularly appealing because the fiber patterns are embedded internally, allowing the design of a robot without modifying its overall form factor.
    We created six expressions for \textit{Sprout} using four types of deformations. 
    Through online and in-person user studies, we investigated how users interpret \textit{Sprout}’s expressions; how they perceive \textit{Sprout} as a social robot; and what design requirements and use cases designers and engineers imagine for \textit{Sprout}. 

    In the remainder of the paper, we outline relevant work and describe our design process. We present the procedures and findings of our user studies. The paper concludes with a discussion of the design implications and limitations of our work. Our work makes the following contributions:
    \begin{enumerate}
        \item \textit{Design}: a novel design space that uses \textit{fiber-embedded actuators} to construct and animate robot bodies;
        \item \textit{Artifact}: \textit{Sprout} as a novel robot design that extends the design space of expressive social robots;
        \item \textit{Empirical}: an understanding of how people perceive \textit{Sprout}'s expressions, the robot as a character, and its potential use.
    \end{enumerate}

\begin{figure*}[!t]
    \centering
    \includegraphics[width=\linewidth]{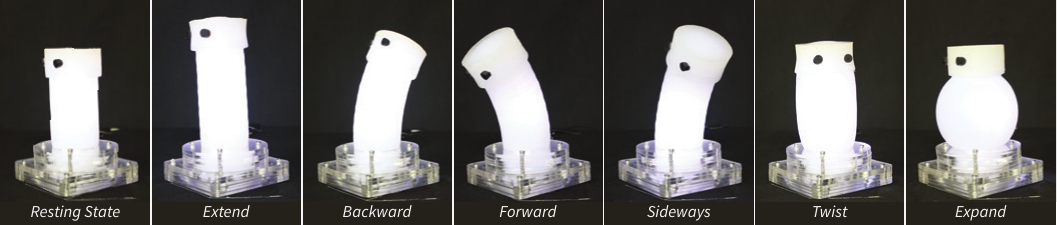}
    \caption{\modified{Still images from the videos that displayed \textit{Sprout}'s resting state and six motion patterns. 
    }}
    \Description[]{Six photos are aligned in one row. All photos include a backdrop on a Sprout at a front view.}
    \label{fig:sixmotions}
\end{figure*}

\section{Related Work}
    Our work builds on literature on (1) novel robot social cues in HRI, (2) prior explorations of soft materials in HRI design, and (3) interaction design using shape-changing interfaces.

\subsection{Novel Robot Social Cues}

    Previous HRI research focused primarily on enabling robots to mimic human communication modalities, such as facial expressions and gestures, for effective interaction. \modified{Although this approach might be appropriate for humanlike robot designs, there is a need to identify new forms of and modalities for expression to meet the needs and goals of diverse robot designs.}

    Prior work has leveraged Augmented and Mixed Reality (AR/MR) technology to project information onto robots, attached physical devices to robots, and introduced new materials to enhance the communicative abilities of robots. 
    For example, AR/MR techniques can be employed to project signals that convey the motion intent of a flying robot \cite{Walker2018}. 
    For a robot that lacks physical expressivity, virtual arms can be added to enable engagement with users  \cite{Groechel2019}.    
    The physical augmentation approach encompasses a variety of communication modalities, including lights \cite{Daniel2015LED, Cha2017, Song2018}, fluid \cite{Koike2023, Guo2020}, and physical attachments \cite{Singh2013Dogtail, singh2013doghci, Hooman2020}. 

\subsection{Use of Soft Materials for Robot Design} %

    Prior work in HRI has investigated how incorporating soft materials into social robot designs can evoke warmth \cite{DiSalvo2003} as well as foster attachment and provide affective touch experiences \cite[]{Sato2020, Yohanan2011}. \citet{DiSalvo2003} developed a robotic product designed to facilitate intimate communication over long distances. The robot is covered with velour and silk upholstered fabrics, seamlessly integrating into domestic environments. \citet{Sato2020} proposed affective touch robots that combine textures and movements to enable richer emotional expression. 
    \citet{Yohanan2011} explored creating affective touch experiences by developing animal-like robots. Meanwhile, \citet{Stiehl2005} introduced a teddy bear-style robot covered in fur fabric to create a robot companion akin to therapy animals. 
    In these works, soft materials were employed as an indirect design element in support of the goals or context of the interaction. 
    In more practical terms, creating safe human-robot interaction is another reason to use soft materials. For example, \citet{Kim2015} developed \modified{a soft skin module using rubber-like flexible material, and \citet{Alspach2015} used the skin module to construct a robot that is safe and robust to support physical interactions between the robot and children.}

    Recently, several HRI studies explored the creation of novel robot expressions using silicone-based materials \cite{Hu2020, Sabinson2021, sabinson2021pheb, klausen2022signalling, farhadi2022exploring, larsen2022wisard}. \citet{Hu2020} developed \modified{a pneumatically actuated skin for robots expressing emotions by changing the robot's skin (\textit{i.e.}, goosebumps and spikes), which can serve as both} visual and haptic cues. \citet{Sabinson2021} introduced a pneumatic-actuated robotic surface to assist in regulating users' emotional states in confined spaces.
    \citet{klausen2022signalling} designed a non-anthropomorphic robot using silicone-based material that communicates emotions through breathing patterns. \citet{larsen2022wisard} implemented a dermis for a robot arm inflating to provide weight information.    
    
\subsection{Shape-Changing Interfaces} 
    Our work is also informed by the literature on shape-changing interfaces within the field of human-computer interaction (HCI). A shape-changing interface is capable of physically transforming or changing its form, thereby enabling novel forms of interaction between humans and machines. 
    \citet{Lining2013pneui} proposed a programmable shape-changing interface and explored its design space, including a height-changing tangible icon for a mobile app or a transformable tablet case. \citet{Cui2023} presented a construction kit for a shape-changing interface and proposed designing robot behavior as its application. 
    \modified{Prior work has also explored how shape-changing interfaces can communicate emotional states} \cite[]{Park2014wrigglo, Tan2016, Dawson2013, Davis2015}. 
    For instance, \citet{Park2014wrigglo} explored how a shape-changing attachment for smartphones can allow users to share physical movements.
    \citet{Tan2016} created a shape-changing interface expressing emotional states, including six emotions (\textit{i.e.,} happiness, sadness, anger, surprise, fear, and disgust).    
    While these projects showcase the expressive potential of shape-changing actuators, \modified{little is known about how they might be designed to support human-robot interaction and to convey diverse emotional states.}

\section{Design and Implementation}


\subsection{Design Inspiration}
    
    In this project, we are originally motivated by two key questions: \textit{How can soft materials be used to enhance robot expressiveness?} and \textit{What would a soft expressive robot look like?} 
    To tackle these questions, we brainstormed what kind of expressions rigid-bodied robots had not achieved yet. During our exploration of ideas, we found our inspiration from the world of animation. In animation, characters sometimes convey their emotions or internal states through exaggerated body deformations. 
    For example, an inflated belly can show that the character ate too much, or an extended neck can express their curiosity. Inspired by these observations, we conceptualized a robot that can change its body shape. 

\subsection{Fiber-Embedded Actuators}

    To bring our concept to life, we leverage \textit{fiber-embedded actuators} to express animation-like robot bodily expressions \cite{Connolly2016}. 
    The \textit{fiber-embedded actuator} takes the form of a hollow elastomeric cylinder with fiber-embedded walls.     
    If a cylinder with no embedded fibers were inflated, it would expand uniformly like an inflated balloon. When flexible but inextensible fibers are embedded in the walls of the cylinder, they limit Sprout’s deformation in the direction of the fibers’ axes. By utilizing different types of fiber patterns, we can achieve various deformations.
    
    In this work, we explored four types of deformation: extension, bending, twisting, and expansion. 
    To achieve extension, we use a layup pattern of circles. This constrains radial expansion while allowing lengthening. 
    To achieve a bending motion, we begin with the circular layup described in the extension motif, but we add reinforcing fibers on one side of the device to limit extension on that side. 
    To achieve a twisting motion, we guide the deformation direction by wrapping the fibers in a spiral pattern. Upon inflation, the spiral fibers cannot extend, thus they cause overall twisting of the actuator in the opposite direction of the fiber spiral. 
    To achieve expansion without lengthening, we align the fibers vertically to limit extension. 
    Figure \ref{fig:teaser} (a) visually represents the fiber patterns.
            
\subsection{Designing Sprout}    
     
    \modified{
    Our design of \textit{Sprout}'s expressions was inspired by the expressions generated by the \textit{half-filled flour sack}, introduced in illustrating the \textit{Stretch and Squash} animation principle \cite{thomas1981illusion}.
    This principle is based on the observation that any object, unless it is stiff, changes its shape during action but maintains its volume.}
    \modified{\citet{thomas1981illusion} argued that this principle can enable animators to ``strengthen characters' actions,'' and create ``a feeling of moving flesh.''
    In their illustrations of this principle, the \textit{half-filled flour sack} took many poses, including squashed, extended, twisted, or bent, by distributing the flour in the sack while maintaining volume.}
    \modified{
    We find inspiration from the rich expressivity that can be achieved by changing the shape of such a simple form, \textit{e.g.,} bending the flour sack forward to express dejection or stretching it to express joy.}
    \modified{
    With this inspiration, we created six expressions, namely \textit{extend}, \textit{backward}, \textit{forward}, \textit{sideways}, \textit{twist}, and \textit{expand} (Figure \ref{fig:sixmotions}).}

    \textit{Sprout} itself features a simple design, consisting of a head with two eyes and a cylindrical body. This simplicity in form was intentional, allowing us to focus on the expressions and minimize the influence of other factors on study participants. 


\subsection{Construction of Sprout}

    \textit{Sprout} is fabricated using a sequential mold-casting technique. We use an elastomer (Ecoflex 00--30 \footnote{\url{https://www.smooth-on.com/products/ecoflex-00-30/}}) for the body material, and Kevlar threads as inextensible reinforcing fibers. All molds are 3D printed with an SLA 3D printer. 
    The fabrication process for \textit{Sprout}'s body can be summarized as follows: 1) fabricating the inner-elastomeric layer, 2) laying up a fiber pattern on the outside of the inner-elastomeric layer, 3) fabricating the outer-elastomeric layer, encapsulating the fiber and inner-elastomeric layers. 
    The head of \textit{Sprout} is fabricated of the same elastomer in a separate mold, then integrated with the body. 
    Once the body and head are bonded, the tube and eyes are added, and then \textit{Sprout} is secured in an acrylic stand. 
    \textit{Sprout} stands at a height of $100 mm$, with a body diameter of $50 mm$.
    
    To animate \textit{Sprout}, we employed a pneumatic system with an electronic pressure regulator controlled using a microcontroller.
    The pressure levels to yield the desired pose for each expression are $20.0 kPa$ for \textit{extend}; $27.5 kPa$ for \textit{backward, forward, and sideways}; $20.7 kPa$ for \textit{twist}; and $22.5 kPa$ for \textit{expand}. 

    \modified{A comprehensive description of the fabrication and characterization is provided by \citet{Koike2023_2} in a separate publication.}

\section{User Study}

    To understand the expressive capabilities of \textit{Sprout}, we conducted two evaluations: an online video study and an in-person feedback session.
    \modified{
    The online study focused on gathering overall user perceptions of \textit{Sprout}, such as perceived emotion of \textit{Sprout}'s expressions, while the in-person session solicited input from designers and engineers to identify directions for further exploration.} 
    \modified{All study activities were reviewed and approved by the University of Wisconsin–Madison Institutional Review Board (IRB)}.\footnote{Questionnaires and videos used for the studies are available at \url{https://osf.io/7u2e6/}}

\subsection{Study 1: Online Video Study}

    \subsubsection{Participants} 
    We recruited 100 participants (47 male, 51 female, 2 other), aged 18--63 ($M = 32.00$, $SD = 12.01$), through an online study platform called \textit{Prolific}\footnote{\url{https://www.prolific.co/}}. 
    All participants were geographically located in the United States and were fluent English speakers. The entire procedure took participants 20 minutes to complete, and they received \$4 USD as compensation. 
    \modified{We excluded five low-quality responses (indicated by, \textit{e.g.,} lack of response to study prompts) and used 95 responses for our analysis.}
    
    \subsubsection{Study Design}  
    We presented six brief videos, each corresponding to different expressions of \textit{Sprout}: \textit{extend}, \textit{backward}, \textit{forward}, \textit{sideways}, \textit{twist}, and \textit{expand}.
    Each video presented \textit{Sprout} from the front and at a 45$^{\circ}$ angle. The duration of each video was approximately five seconds. Participants were shown the videos in a randomized order. 
    After each video was presented, participants were asked to select emotion words from an emotion wheel (Figure \ref{fig:emotionrank} (a)) \cite{plutchik1980general, plutchik1988nature} that they believed best described the emotions conveyed by \textit{Sprout}. If the available options on the emotion wheel were insufficient in capturing their perceptions, participants were encouraged to provide additional words through a free-form entry. 
    Following the emotion labeling tasks, participants were asked to suggest situations in which \textit{Sprout's} emotions might be appropriate for their selected emotion words.
    \modified{
    Finally, participants rated their overall perceptions of the robot using the Godspeed Questionnaire \cite{Bartneck2009MeasurementIF} on a five-point rating scale.}

\subsection{Study 2: In-person Feedback Sessions}

    \subsubsection{Participants} 
    We recruited 13 participants (9 male, 4 female), aged 21--67 ($M = 30.00$, $SD = 13.79$), for in-person feedback sessions via a campus mailing list.    
    To solicit insightful feedback on our current designs of \textit{Sprout} and explore potential improvements, we recruited participants with backgrounds in art, design, and mechanical engineering. All participants were geographically located in the United States and were fluent English speakers. The entire procedure took participants 30 minutes, and they received \$8 USD.
    
    \subsubsection{Study Design}
    
    At the beginning of each session, the experimenter provided an explanation of the study's purpose as ``exploring design possibilities using soft robotic techniques for creating new robot expressions,'' and demonstrated \modified{four expressions of \textit{Sprout}: \textit{extend}, \textit{forward}, \textit{twist}, and \textit{expand}.
    We did not include \textit{backward} and \textit{sideways} due to time constraints but verbally explained that the bend deformation can generate those two expressions.}
    Following the introduction, participants engaged in a design feedback session that lasted 15--20 minutes. In the session, they were asked semi-structured questions such as ``What were the use cases that came to your mind when you first saw this robot,'' or ``How do you think we can improve the design of Sprout?'' Additionally, we provided participants with craft materials to facilitate discussions around future use cases and design suggestions for \textit{Sprout}. Figure \ref{fig:inperson} shows the scenes from the in-person study.

\begin{figure*}[!b]
    \centering
    \includegraphics[width=\linewidth]{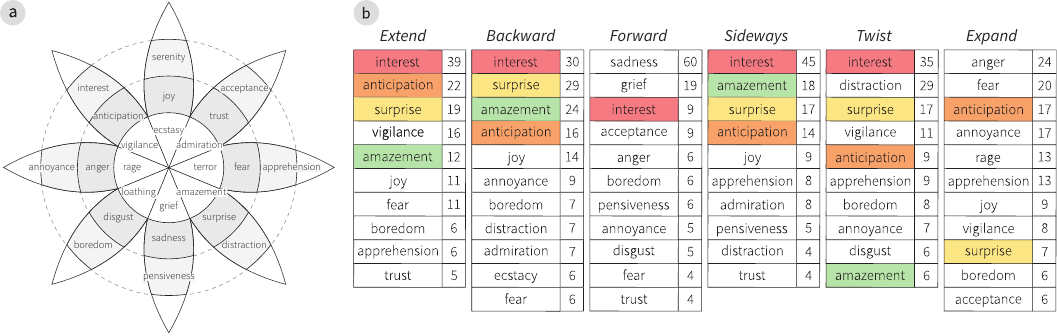}
    \caption{(a) Plutchik's Emotion Wheel \cite{plutchik1980general, plutchik1988nature}  used for the emotion labeling task, and (b) results from the task for each expression. The columns show the Top-10 emotion words chosen for each expression.
    The highlighted boxes correspond to emotion words that received a high score (\textit{i.e.,} top five within an expression) in at least three expressions. 
    Each highlighted emotion word is associated with the same color across expressions.}
    \Description{An illustration is on the left. A set of tables is on the right. An illustration shows Plutchik’s emotion Wheel in a grayscale. A set of tables on the right shows the results from the emotion labeling task. There are 6 tables in total, each table has two columns and 10 to 12 rows.}
    \label{fig:emotionrank}
\end{figure*}

\section{Results}

\subsection{Study 1: Online Video-based Study}

\subsubsection{Emotion Labeling through Plutchik's Emotion Wheel} \label{sec:results:plutchik}
    We analyzed and visualized participant \modified{emotion-labeling} responses through Plutchik's emotion wheel in Figure \ref{fig:emotionrank} (b). 

    \textit{Interest} was the most frequently used, ranking highest for \textit{extend}, \textit{backward}, \textit{sideways}, and \textit{twist}, and the third highest for \textit{forward}. \textit{Surprise} was also frequently used, ranking second in \textit{backward} and third in \textit{extend}, \textit{sideways}, and \textit{twist}. Finally, \textit{anticipation} ranked in the top five for \textit{extend}, \textit{backward}, \textit{sideways}, \textit{twist}, and \textit{expand}.

    \textit{Extend}, \textit{backward}, and \textit{sideways} have similar word-choice patterns, all including \textit{interest, surprise, amazement,} and \textit{anticipation} in their top five words. \textit{Twist} also contains the same four words, with an addition of \textit{distraction} that differentiates it from \textit{backward}, \textit{sideways}, and \textit{extend}.
    \textit{Forward} and \textit{expand}, on the other hand, seem to be independent compared to other expressions.  \textit{Forward} was strongly associated with \textit{sadness} and \textit{expand} with \textit{anger}.

    The words from three sets of emotions, ecstasy--joy--serenity, loathing--disgust--boredom, and admiration--trust--acceptance, were not frequently used for tagging across the six expressions.

\subsubsection{Emotion Labeling through Free Entry} \label{sec:results:emotionfree}
    Table \ref{table:openendedemotions} summarizes the results from the free-entry emotion-labeling task. We found that  \textit{curiosity} was used for all expressions except \textit{expand}. \textit{Sideways} received the highest number of mentions, followed by \textit{extend}, \textit{twist}, \textit{forward}, and \textit{backward}. 
    Apart from curiosity,
    \textit{extend} received words indicating attention of \textit{Sprout}, such as \textit{attentiveness}, \textit{fascinated}, and \textit{watchful}.
    Responses for \textit{backward} encompassed various emotional states, some conflicting with each other. It included words describing unpleasant feelings like \textit{discomfort} and \textit{exhaustion}, as well as pleasant-feeling words like \textit{relief}.
    \textit{Forward} received mixed responses, including words related to courtesy such as \textit{respect} and \textit{politeness}, as well as negative feelings like \textit{ashamed} and \textit{disappointment}.
    Responses for \textit{sideways} included words indicating uncertainty, such as \textit{confusion} and \textit{questioning}.
    For \textit{twist}, responses included unpleasant emotions such as \textit{rejection}, and words related to heightened awareness, like \textit{alertness}.
    Participants described \textit{expand} with words such as \textit{defensive} and \textit{intimidation}. Additionally, it obtained words like \textit{overeating} that indicate bodily conditions rather than emotions.

\subsubsection{Situation Description}

    \modified{
    We analyzed 570 responses (95 per expression) in the situation-description task. The first author and an external coder, hereinafter ``coders,'' conducted a thematic analysis of the responses using an inductive (data-driven) approach. 
    The coders independently analyzed two complete sets, consolidating codes into categories: 
    (1) object or actor \textit{Sprout} reacts to, (2) action describing \textit{Sprout}'s emotional state, and (3) gesture or metaphorical language explaining \textit{Sprout}'s behavior. The coders then individually coded the remaining four expressions using the three aforementioned categories. After all responses were coded, all codes were reviewed, addressing disagreements through re-coding.}   
    \modified{
    Finally, we iteratively clustered the codes through affinity diagramming to identify contexts where \textit{Sprout} can be found, how people imagine \textit{Sprout}'s participation in a situation, or the nuances of \textit{Sprout}'s emotional states that could not be observed in the emotion-labeling task (\S \ref{sec:results:plutchik} and \ref{sec:results:emotionfree}).} 
    \modified{
    In reporting our results below, the number in parentheses after an italicized category refers to the number of codes clustered into the category.
    }

    \modified{
    \textit{Extend.} 
        Participants described \textit{extend} as a response to \textit{potential danger} (17) for \textit{Sprout}, such as ``a threat'' or ``something unexpected.'' 
        Some participants explained \textit{Sprout}'s response as \textit{a reaction to a user} (17) during a conversation, \textit{e.g.,} question, command, or greeting.
        \textit{Extend} was explained with languages indicating \textit{Sprout}'s \textit{active curiosity} (32), \textit{e.g.,} ``observe keenly'' and ``perking up.''}
    
    \modified{
    \textit{Backward.} 
        Participants described \textit{backward} as a reaction to \textit{something from above} (19), \textit{e.g.,} someone taller than \textit{Sprout}, the sky, or a scary monster. Participants interpreted \textit{backward} as \textit{a pleasant reaction} (12) in scenarios such as receiving ``a present,'' or being surrounded with ``calming ambience or music.'' 
        \textit{Backward} was seen as a form of \textit{ruminating behavior} (21), such as ``looking up'' and ``investigating.''
        \textit{Backward} was also seen as \textit{unpleasant behavior} (14), \textit{e.g.,} ``irritation'' and ``frustrated sigh.''
        Participants expressed that \textit{Sprout} had \textit{humanlike expressions of disgust} (\textit{e.g.,} ``rolling its eyes.'')}

    \modified{
    \textit{Forward.} 
        Participants described \textit{forward} in situations where \textit{Sprout} received or experienced \textit{unpleasant information or event} (28), such as ``bad news,'' ``being scolded,'' or ``loss of someone important.''
        On the other hand, some participants interpreted \textit{forward} as a form of \textit{acknowledgment} (15), \textit{e.g., }``bowing'' to show respect or apology, or ``nodding'' to express agreement or sympathy.}

    \modified{
    \textit{Sideways.}
        \textit{Sideways} was seen as a reaction when \textit{Sprout} ran into \textit{something unfamiliar} (16), \textit{e.g.,} ``a question that it doesn't know the answer to.''
        Some participants perceived \textit{Sprout} as \textit{waiting} (15), \textit{e.g.,} ``waiting for [a] reply,'' or ``processing'' information.
        We note that 11 participants compared \textit{forward} to a human's or dog's behavior of tilting their head ``when they are curious.'' }
    
    \modified{
    \textit{Twist.}
        Participants described \textit{twist} as a reaction to \textit{unexpected objects} (25) appearing or sounds heard near the robot's side, such as ``a noise'' or ``something moving by.'' Participants also interpreted \textit{twist} as a reaction to \textit{the presence of others in the same room} (11), \textit{e.g.,} either of someone who ``walked into the room'' or ``[left] the room.''}

    \modified{
    \textit{Expand.} 
        Participants described \textit{expand} as a reaction to \textit{unpleasant stimuli} such as ``a person it doesn't trust,'' or someone who ``was mean to'' \textit{Sprout}, namely \textit{a response to a threat} (29) by ``making [itself] appear bigger'' or ``taking a deep breath'' to ``remain calm.''
        Some participants compared \textit{expand} to \textit{a defensive mechanism in animals} (16), such as a puffer fish that inflates its body when it is in danger. 
        Two participants included cartoon expressions, \textit{e.g.,} ``a cartoon character with steam exuding from [its] ears.''}

\subsubsection{GodSpeed Questionnaire} 

\begin{figure*}[!t]
    \centering
    \includegraphics[width=\linewidth]{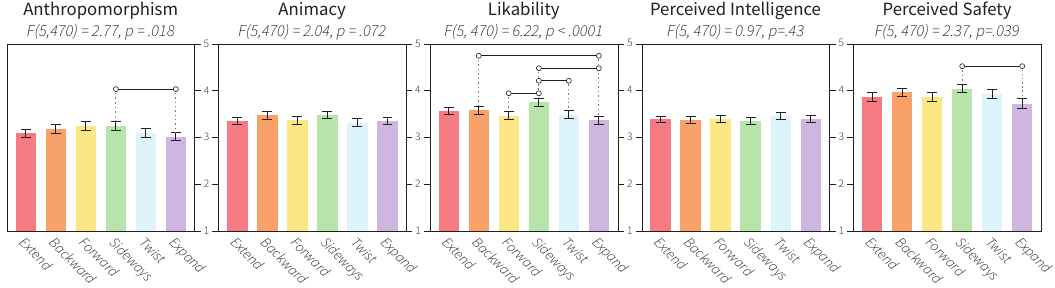}
    \caption{\modified{Results from the GodSpeed Questionnaire data across six expressions for Anthropomorphism, Animacy, Likability, Perceived Intelligence, and Perceived Safety.}}
    \Description{Six box plots are aligned in one row. Each box plot has a subscale of the GodSpeed questionnaire. Each box plot has six bars assigned for six expressions, respectively. The bars indicate extend, backward, forward, sideways, twist, and expand, from left to right.}
    \label{fig:godspeed}
\end{figure*}

    \modified{Our analysis of the data from the Godspeed Questionnaire started with calculating the reliability of its subscales (Anthropomorphism: $\alpha = .867$; Animacy: $\alpha = .856$; Likeability: $\alpha = .916$; Perceived Intelligence: $\alpha = .891$; Perceived Safety: $\alpha = .876$ ).}
    \modified{
    To analyze data from each subscale, we conducted an analysis of covariance (ANCOVA), using the subscale as the output variable, the robot expression (\textit{extend}, \textit{backward}, \textit{forward}, \textit{sideways}, \textit{twist}, \textit{expand}) as the input variable, and participant age and gender as covariates to control for these factors when they had a significant effect on the output variable. }
    \modified{
    Robot expression had a significant effect on Anthropomorphism ($\textit{F}(5,470) = 2.77, \textit{p} = .018$), controlling for age ($\textit{p} = .031$); a marginal effect on Animacy ($\textit{F}(5,470) = 2.04, \textit{p} = .072$) controlling for age ($\textit{p} = .022$); a significant effect on Likability ($\textit{F}(5, 470) = 6.22, \textit{p} < .001$); and a significant effect on Safety ($\textit{F}(5, 470) = 2.37, \textit{p} =.039$); but no significant effect on Perceived Intelligence ($\textit{F}(5,470) = 0.971, \textit{p} = .435$). }
    \modified{
    For each subscale where robot expression had a significant effect, we calculated pairwise comparisons using Tukey's Honestly Significant Test (HSD), which showed that \textit{sideways} was seen as being more anthropomorphic, more likable, and safer than \textit{expand}. \textit{Sideways} was also more likeable than \textit{forward} and \textit{twist}; and \textit{backward} was more likeable than \textit{expand}.}
    Figure \ref{fig:godspeed} summarizes these results. 

\begin{table}[!b]
    \caption{Results from the emotion labeling task through the free-entry form. The numbers in parentheses after each word indicate the frequency of use by participants.}
    \Description{The list of open-ended responses across six expressions is shown in the table with six rows and two columns.}
    \label{table:openendedemotions}
    \centering
    \small
    \begin{tabular}{cl}
    \toprule
        \textit{Extend} & \textbf{Curiosity (4)}; Attentiveness (2); Comfortable; Fascinated; \\
        & Irritation; Watchful; Friendliness; Indifferent\\ 
    \midrule
        \textit{Backward} & Laughing (2); Relief (2); Calm; \textbf{Curiosity}; Disagreement; \\
        & Discomfort; Exhaustion; Realization; Uninterest \\
    \midrule
        \textit{Forward} & Ashamed (2); \textbf{Curiosity (2)};  Disappointment (2); Upset; \\
        & Respect (2); Greeting; Intimidated; Politeness; Sorrow; \\
    \midrule
        \textit{Sideways} & \textbf{Curiosity (18)}; Confusion (14); Questioning (3); \\
        & Attentive; Bewildered; Disbelief; Happiness; Intrigue; \\
        & Leap with Joy; Puzzlement \\
    \midrule
        \textit{Twist} & \textbf{Curiosity (3)}; Alertness (2); Rejection (2); Apathy; \\
        & Disgust; Observant; Pride; Sarcasm; Startled \\
    \midrule
        & Overeating (5);  Defensive (3); Intimidation (3); \\
        \textit{Expand} & Happiness (2); Calm; Deep Breath; Depressed; \\
        & Holding breath; Laziness; Love; Overwhelmed; Pride; Sick \\
    \bottomrule
\end{tabular}
\end{table}

\subsection{Study 2: In-Person Feedback}

    \modified{
    The first author and an external coder, hereinafter ``coders,'' conducted a thematic analysis of transcriptions of 13 interview recordings. The primary coder coded and categorized transcripts, resulting in 33 codes and four main categories. The secondary coder reviewed all transcripts, addressing disagreements with the primary coder through re-coding and re-categorization.}
    Our analysis revealed four themes: (1) \textit{Sprout}'s expressions, (2) perceptions of \textit{Sprout}, (3) design recommendations to improve \textit{Sprout}, and (4) suggestions for \textit{Sprout}'s use cases. For each theme, we describe how descriptions differed across deformation and provide illustrative quotes from our data. The quotes are marked with participant IDs (\textit{e.g.}, P2 denotes ``participant 2'').

\subsubsection{\textit{Sprout}'s Expressions} 

    Participants provided commentary on what they thought \textit{Sprout} expressed through the four expressions they observed, which we describe below.

    \textit{Extend.}
    P5 and P10 mentioned that \textit{extend}'s up and down movement can express ``excitement.'' P6 saw this movement as expressing ``it's happy to see me'' when the participant wakes up in the morning. P7 and P11 mentioned that \textit{extend} can be used ``[to show] power or [project] confidence'' because its movement of getting larger is ``intimidating.''
    Four participants (P4, P7, P12, P13), however, did not see emotions in \textit{extend}, \textit{e.g,} P12 said, ``It's just going up and down.'' This movement instead reminded them of worms or spotted garden eels popping out of the sand.

    \textit{Forward.}
    P6 and P8 mentioned that they felt uncomfortable when they saw the \textit{forward}. P8 mentioned \textit{forward} reminded them of a ``weird bird neck.'' On the other hand, P4 felt \textit{forward} is cute because they imagined \textit{forward} is like playing peek-a-boo: ``If it's behind the wall, it's like peeking out.'' Furthermore, five participants (P2, P4, P7, P10, P12) had a positive impression towards \textit{forward} because they described \textit{forward} as a polite greeting to them. 

    \textit{Twist.}
    P1 and P3 implied that \textit{twist} lacks the capability of expressing emotion. Four participants (P4, P7, P8, P9) mentioned that \textit{twist} can indicate directions or get people's attention; P8 said, ``[\textit{Twist}] could also get people to look [in] a certain direction.'' Furthermore, P8 said \textit{twist} looks like \textit{Sprout} is lost.
    
    \textit{Expand.}
    \textit{Expand} was chosen as the first favorite of four participants (P1, P8, P11, P13). \textit{Expand} was described with positive words such as ``cute'' (P1, P13), ``humorous'' (P7), or ``goofy'' (P4). P2 and P4 mentioned \textit{expand} can ``make children laugh.'' P2 and P4 puffed out their cheeks to demonstrate how to make children laugh and shared that \textit{expand} reminded them of that.
    Furthermore, \textit{expand} was seen as expressing ``overeating'' (P2, P5) and ``breathing'' (P1, P8, P9, P11). 
    Three participants (P1, P3, P10) mentioned \textit{expand} can express ``anger,'' ``frustration,'' or ``stress.'' 
    P1 imagined \textit{expand} used in a situation where an education assistant is in a math class: ``If there's a hard problem, \textit{expand} kind of just shows it's stressful.''

\subsubsection{Perceptions of Sprout}
\begin{figure*}[!t]
    \vspace{-2mm}
    \centering
    \includegraphics[width=\linewidth]{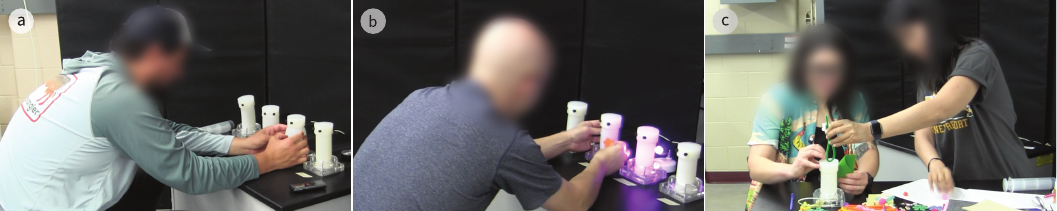}
    \caption{Scenes from the feedback session. (a) P5 touching \textit{Sprout} to understand how it felt when it was moving. (b) P9 exploring how LED light might diffuse through \textit{Sprout}'s body. (c) P11 and an experimenter exploring a new character design for \textit{Sprout}. }
    \Description{Three photos are aligned. In the leftmost photo, a participant is sitting on a chair and holding a Sprout with both hands. In the photo in the center, a participant uses a spotlight with one hand and the other holds Sprout. In the rightmost figure, two people are working together. A person on the left is a participant and the one on the right is an experimenter. They are putting crafting material on the top of Sprout.}
    \label{fig:inperson}
    \vspace{-2mm}
\end{figure*}

    All participants had a positive impression of \textit{Sprout}: \textit{Sprout} was referred to as ``cute'' (P5) and ``fun to watch'' (P9). Additionally, seven participants used animal references, likening \textit{Sprout}'s behavior to that of dogs, cats, snakes, or spotted garden eels. Furthermore, three participants (P10, P11, P13) mentioned that \textit{Sprout} reminded them of animated characters such as Baymax (P13), and Pokémon (P11). Four participants (P1, P2, P5, P6) expressed their desire for more physical interaction with \textit{Sprout}. P2 envisioned \textit{Sprout} pushing back when users poke it. P5 touched \textit{Sprout} to understand how it feels when it is in motion (see Figure \ref{fig:inperson}(a)). In addition, five participants (P2, P3, P5, P6, P13) want \textit{Sprout} to sense and react to users' behaviors.

\subsubsection{Design Recommendations to Improve \textit{Sprout}}
    Participants' suggestions for \textit{Sprout} can be categorized into two areas: designing motion and adding other modalities. 
    For designing motion, participants suggested that different speeds (frequencies) can create various impressions or convey different meanings. Factors such as pressure (amplitude), combinations of deformation patterns within a single body, and the use of ease in/ease out were also mentioned as potential parameters for motion design. 
    As for adding other modalities, participants mentioned adding color and sound. For instance, P4 suggested humorous sounds to exaggerate the deformations, and they mentioned that \textit{Sprout} could be more expressive by communicating with users through beeping, similar to the R2D2 robot from the Star Wars franchise. As an example of using colors with \textit{Sprout}, P9 proposed fading the body color using LED lights in conjunction with body deformation.

    In addition to enhancing expressiveness, participants provided suggestions related to modifying \textit{Sprout}'s physical appearance to make it more appealing. For instance, they suggested improving the shape of \textit{Sprout} by updating its head to ``a rounded head'' (P7) or more ``facial features'' (P3), such as googly eyes (P1), a mouth and a nose (P2), blushed cheeks (P11) or emoji-like facial expressions (P3). P8 and P11 suggested a new character (robot) design using the same deformation patterns, with P8 sketching a mushroom robot. Furthermore, some participants expressed concerns about the material, commenting on silicone discoloration, sustainability, and dustiness.   
    
\subsubsection{Suggestions for \textit{Sprout}'s Use Cases}
    The use cases that emerged from the participants can be categorized into five contexts: healthcare, media (\textit{e.g.,} for interactive art installations), conversational agents, home environments, and children-centered contexts, which include educational settings and childcare.
    
    In a \textit{healthcare} context, P9 and P11 mentioned that \textit{Sprout} can assist with breathing and mindfulness exercises, given its breathing-like motion. P3 also discussed the potential to use \textit{Sprout} to help individuals with social deficits better express themselves. 
    
    \textit{Sprout} can also serve as a delightful \textit{interactive display} at airports, museums, or tables. P11 suggested that \textit{Sprout} could ``grow with the sunlight'' to indicate the passage of time, and they also envisioned multiple \textit{Sprouts} on a desk interacting with each other.
   
    P13 envisioned \textit{Sprout} as a \textit{conversational partner} that users can engage with and receive reactions from. Several participants noted that repetitive \textit{forward} expressions could convey ``nodding'' and facilitate dialog.
    
    Within a \textit{home} environment, many participants likened \textit{Sprout} to home assistant devices like Google Home. \textit{Sprout} could function as a more expressive household robot similar to these devices.
    
    Seven participants suggested that \textit{Sprout} could serve as an \textit{educational assistant}. For example, P1 proposed that ``\textit{Sprout} can teach music by moving to the beat,'' while others thought \textit{Sprout} could \textit{expand} to express stress during challenging tasks in a classroom. In addition to educational roles, \textit{Sprout} could be used for \textit{younger children}, even infants. P4 said that \textit{Sprout} should be able to sense when a baby is crying and make the baby laugh.


\section{Discussion}
 
\subsection{Interpretation of Each Expression} \label{sec:discussion:interpretation}

    Overall, findings from the two studies suggest that each expression is capable of conveying particular emotional or internal states. 
        
    \modified{
    \textit{Extend} can demonstrate \textit{active engagement} with a situation or interaction with users. It can particularly express \textit{interest} or \textit{excitement} toward something unknown to \textit{Sprout}, or something that arouses its curiosity.
    \textit{Extend} can also communicate \textit{Sprout's} presence to users, \textit{e.g.,} greeting or responding to a command. }

    \modified{
    \textit{Backward} can express \textit{surprise} to an unexpected event or \textit{amazement} toward pleasant stimuli. It can also express internal states when \textit{Sprout} is \textit{investigating} or \textit{processing} a situation. While aforementioned emotional states range from positive to neutral, \textit{backward} can also display \textit{Sprout}'s unpleasant states, such as annoyed or exhausted, as its motion resembles the gesture of eye-rolling.}
    
    \modified{
    \textit{Forward} is one of the most distinct expressions that can show negative affect, including \textit{sadness} or \textit{grief}. Compared to other expressions, \textit{Sprout} can show rather passive behavior, \textit{e.g.,} showing sympathy when being told sad news, or feeling smaller when being scolded. Additionally, \textit{forward} can show \textit{acknowledgment} and \textit{respect,} because it has associations with nodding and bowing.}

   \modified{
   \textit{Sideways} can show \textit{Sprout}'s internal states when it faces something it is \textit{unsure} or \textit{uncertain} about, \textit{e.g.}, showing \textit{Sprout} is thinking of an answer to a user's question. It is similar to \textit{extend} and \textit{backward} in displaying \textit{interest} toward something unexpected. \textit{Sideways} may be a more natural choice as users associated \textit{sideways} with humans and dogs tilting their heads when questioning something.}     
    
   \modified{
   \textit{Twist} can be found in several different situations where it would be able to express \textit{attention}, \textit{distraction}, or \textit{rejection}. If Sprout exhibits \textit{twist} in response to a change in the environment, the action can be interpreted as \textit{attention}. On the other hand, if \textit{Sprout} exhibits \textit{twist} during an interaction with users, it can be seen as \textit{distraction} or \textit{rejection}. Furthermore, \textit{twist} can indicate a direction.}

    \modified{
    \textit{Expand} can communicate \textit{anger} and \textit{fear} when dealing with unpleasant situations.
    Furthermore, \textit{expand} can express bodily motions or conditions, \textit{i.e.,} \textit{breathing} or \textit{overeating}.
    Displaying overeating could be particularly interesting, as it often is a cartoonishly exaggerated expression, and could add an element of \textit{humor} to the robot's expressivity.}

\subsection{Design Implications} \label{sec:discussion:implications}

    Findings indicate that \textit{Sprout} creates a gentle and positive impression, with \textit{interest} being the most frequently used term in emotion labeling. This suggests \textit{Sprout} can convey curiosity and a desire to learn, positioning itself as a personal assistant. Additionally, Study 2 showed that \textit{Sprout} reminds participants of animated characters, suggesting it can serve as an attentive character across all the identified use cases, always available to assist users.

    From the findings of Study 2, there are opportunities to enhance \textit{Sprout}'s expressive capabilities with additional modalities. We observed that several participants expressed a desire to integrate LED color changes with \textit{Sprout}'s motions. Light is a modality commonly employed in various robot platforms (\textit{e.g.}, Amazon Echo) and often studied within HRI research \cite[\textit{e.g.},][]{Daniel2015LED, Cha2017, Song2018}. Its integration should create more nuanced expressions for \textit{Sprout}. 
    
    The motion pattern is another important factor in enriching \textit{Sprout}'s expressive capabilities. While we investigated unidirectional and monotonic motion in this study, exploring bidirectional or symmetrical twisting, bending in both directions (side-to-side or backward-forward), expansion, and contraction, can introduce another layer of expressive possibilities. Additionally, the frequency of motions can serve as a design parameter to enhance \textit{Sprout}'s expressivity. Although this study examined one single frequency for each deformation pattern, frequency has been identified as a highly influential parameter \cite{Hu2020}.
    
    Participants also suggested the inclusion of facial expressions or modifications to character design using the same deformation mechanism. Our initial design for \textit{Sprout} aimed for simplicity to avoid influences on its form factor; however, incorporating a shape-changing body with designing the robot's character and personality poses an exciting direction. For instance, the concept of a mushroom robot proposed by P8 evokes associations with the talking flowers from ``Alice in Wonderland,'' which can be perceived as being peculiar but provokes curiosity. \textit{Fiber-embedded actuators} have the potential to create such robots, and even robots that can ``grow.'' As the robot's personality is also significant factor in designing human-robot interaction \cite[\textit{e.g.},][]{Walker2020, Onoda2019, Hoffman2015}, we believe that our work points to a rich design space where the robot's physical and behavioral design is tightly coupled with its personality.

    The findings posit future applications across various use cases for \textit{Sprout}. As a conversational partner, \textit{Sprout} can stimulate dialogue by conveying different forms of \textit{interest}. For instance, \textit{extend} can express \textit{interest} due to being excited; \textit{backward} can express \textit{interest} that results from being surprised; and \textit{sideways} can convey \textit{interest} stemming from being uncertain.  
    Additionally, \textit{twist} serves a dual purpose by indicating both \textit{interest} and direction, making it suitable as a facilitator for turn-taking in multi-user interaction scenarios.
    In healthcare contexts, \textit{expand} was predominantly considered as a partner for breathing exercises. Other expressions, however, can also be utilized in this context. 
    For instance, \textit{forward} can employ a bowing motion to convey a polite greeting, thereby building trust with patients. Similarly, \textit{sideways} can use its motion to inquire, ``How are you doing?''
    In educational settings, repetitive motions can enhance classroom activities, while \textit{twist} can capture students' attention or signal ``Do not do that'' when students lose focus.
    In-home or office environments as an interactive medium, \textit{forward} can subtly communicate messages like ``your plant needs water'' or ``you've been working too much'' by conveying sadness.

\subsection{Limitations \& Future Work} \label{sec:discussion:limitation}

   Our work has several limitations under three categories: lack of motion design, isolated expressions, and study design.
    
    First, our expression design did not incorporate parameters such as frequency or amplitude, which could have enriched our results. As suggested by participants and discussed in prior work \cite[\textit{e.g.,}][]{Hu2020}, these parameters have the potential to create nuanced expressions or different interpretations within the same deformation pattern. For instance, a rapid motion of \textit{backward} could convey surprise, while a slower motion might signify boredom. 
    Additionally, an extensively inflated \textit{expand} could express intense anger, whereas a slight inflation might show stress. 
    Future work should include these parameters to extend our initial exploration of six expressions. 
    
    \modified{We also acknowledge that our current design only allows each robot to perform a single deformation pattern, which may restrict its applicability in complex human-robot interaction scenarios. Our future work will explore a range of deformation patterns by integrating several fiber layup patterns within a single body to accommodate a more sophisticated and diverse set of expressions.}

    Finally, the results may have been influenced by the decontextualized stimuli. We observed that expressions including \textit{extend}, \textit{backward}, \textit{sideways}, and \textit{twist}, were interpreted with a range of emotions. The inclusion of context would have provided a case-by-case interpretation. Given the exploratory study design, we did not implement a comprehensive interaction for \textit{Sprout}; however, Study 2 revealed a desire for interaction with \textit{Sprout} through touch and reactions to users' behavior. Incorporating these suggestions, we aim to further explore and design how \textit{Sprout} interacts with users. 
\section{Conclusion}
    In this paper, we presented \textit{Sprout}, a soft expressive robot. By integrating \textit{fiber-embedded actuators} into its construction, \textit{Sprout} can change the shape of its body to convey its internal states. We designed and implemented six expressions using four deformation patterns for \textit{Sprout}. We conducted an online study ($n=100$) and an in-person feedback session ($n=13$) to evaluate the expressive capability.
    Our results from two types of studies showed that each expression is capable of conveying particular emotional or internal states, and also revealed potential use cases and design improvement for \textit{Sprout}. We suggest that the integration of soft actuators is a novel design space for designing robot expressions.

\begin{acks}
    \modified{We would like to thank Arissa J. Sato for her assistance with the data analysis and the writing of the paper, and Keng-yu Lin and Tessa Luzney for their assistance in constructing \textit{Sprout}.
    This work was made possible by financial support from the \textit{Sheldon B. and Marianne S. Lubar Professorship}, an \textit{H.I. Romnes Faculty Fellowship} and funding from the \textit{Shigeta Education Foundation}.}
\end{acks}

\bibliographystyle{ACM-Reference-Format}
\bibliography{00_Main}

\end{document}